\documentclass[10pt,twocolumn,letterpaper]{article}

\usepackage{cvpr}
\usepackage{times}
\usepackage{epsfig}
\usepackage{graphicx}
\usepackage{amsmath}
\usepackage{amssymb}
\usepackage{graphicx,import}

\usepackage[pagebackref=true,breaklinks=true,letterpaper=true,colorlinks,bookmarks=false]{hyperref}

\cvprfinalcopy 

\def\cvprPaperID{****} 

\begin{document}

\title{Image Recognition Using Scale Recurrent Neural Networks}

\author{Dong-Qing Zhang\\
ImaginationAI LLC\\
{\tt\small dongqing@gmail.com}
}

\maketitle

\begin{abstract}
Convolutional Neural Network(CNN) has been widely used for image recognition with great success. However, there  are a number of limitations of the current CNN based image recognition paradigm. First, the receptive field of CNN is generally fixed, which limits its recognition capacity when the input image is very large. Second, it lacks the computational scalability for dealing with images with different sizes. Third, it is quite different from human visual system for image recognition, which involves both feadforward and recurrent proprocessing. This paper proposes a different paradigm of image recognition, which can take advantages of variable scales of the input images, has more computational scalabilities, and is more similar to image recognition by human visual system. It is based on recurrent neural network (RNN) defined on image scale with an embeded base CNN, which is named Scale Recurrent Neural Network(SRNN). This RNN based approach makes it easier to deal with images with variable sizes, and allows us to borrow existing RNN techniques, such as LSTM and GRU, to further enhance the recognition accuracy. Our experiments show that the recognition accuracy of a base CNN can be significantly boosted using the proposed SRNN models. It also significantly outperforms the scale ensemble  method, which integrate the results of performing CNN to the input image at different scales, although the computational overhead of using SRNN is negligible.

\end{abstract}

\section{Introduction}

Deep Convolutional Neural Network has achieved tremendous success for image recognition, since it won the ImageNet image recognition task in 2012 \cite{alexnet2012krizhevsky}. However, the current CNN-based image recognition paradigm is based on a feedforward neural network architecture, with fixed network depth and receptive field size. Such paradigm has several limitations. 

First, input images are usually resized to a small fixed size(e.g. 224x224) as the network input. The resizing process often results in loss of details for small objects, leading to loss of recognition capacity. Although, it's psossible to use global average pooling to deal with larger size of images than the native resolution of the CNN, the receptive field size of the CNN remains the same. Consequently, although the CNN can "see" more local details, it also loses its capability to capture larger scale spatial structures of objects. The only way to increase receptive field size is to further increases network depth, which, to some point, would make the computation of training or inference intractable. 

Second, it lacks computational scability for large or variable sized images, in the sense that for both large or small images, it has the same computational cost due to the resizing process. It would be more desirable in real application, especially on the mobile and embedded platforms, that for large images we can have certain accuracy-cost tradeoff that can result in better accuracy if more computational resources are used. 

Third, the feedforward network based paradigm is quite different from human's object recognition system. According to the research of neural science, human's visual perception system involves both feedforward processing and recurrent processing \cite{recur2000lamme}. It is also an multi-scale processing mechanism, where the P pathway that involves Midget retinal ganglion cells and Parvocellular cell in LGN is responsible for a fast feedforward process with less details, while the M pathway is a slower feedforward process with more details.  

Partly inspired by the human visual system for object recognition, this paper explores a different paradigm for image recognition from the traditional CNN-based feedforward network. The paradigm is based on a recurrent network with its recurrence relation defined along image scale rather than the traditional temporal or spatial direction. A CNN base network is embedded into the recurrent network as a feature extractor for different scales. The overall network is named Scale Recurrent Neural Network (SRNN), due to its recurrence relation along the scale direction. The input of a SRNN is a list of images resized  with different scales from a single input image. This allows it to process variable-sized images, with large input images usually having more unrolled steps of the SRNN. An important property and advantage of SRNN is that it can both capture fine details of the input image (by the large-sized copies of image) and the large-scale structures of objects (by the small-sized copies). In other word, it has the capability to significantly increase the range of the receptive field of the neural network. Another benefit of SRNN is its computational scability. The inference process starts from a small-scale image copy with smaller computational cost and lower recognition accuracy. And we can achieve higher accuracy on-demand by feeding additional larger-scale image copies. The inference process can stop at any scale according to the computational budget.

The SRNN based paradigm can be viewed as CNN classifier with scale integration. This scale integration process can also be achieved by a straightforward ensemble classifier of averaging inference results of the base CNN on different image scales. It maybe be also achieved by fully connected layers. However, the recurrent network approach allows training and recognition on images with variable sizes, and it also allows us to borrow existing RNN techniques, such as GRU, for further performance boost. In our experiments, we show that SRNN is able to significantly outperform the simple scale ensemble approach. In addition, our proposed GRU variant for SRNN, called half-GRU, also significantly outperform SRNN witih vanilla RNN.

\section{Related Work}

Multi-scale processing has been pervasively used in image processing and computer vision. For example, the scale space framework is widely used in computer vision, which is based on the assumption that the ideal scale for object recognition or processing is generally unknown a priori, therefore we should treat different scales equally in processing or recognition. This paradigm results in some very successful computer vision algorithms, such as the SIFT detector \cite{sift2004lowe}. 

The multi-scale processing framework used in computer vision may be also partly inspired by human's retina ganglion cells, where the Midget cells have small sizes of dendritic trees that corresponds to fine-scale perception, whereas the parasol cells have larger sizes of dendritic trees  that results in courser-level of perception. 

In the deep learning research area, multi-scale processing has been also actively researched. The spatial pyramid pooling (SPP) method \cite{pyramidpool2014he} attemtps to leverage differnet scales to further enhance the recognition accuracy. It uses a pyramid pooling layer to pool the output feature planes from base network with different scales, and the pooled feature vectors are concatnated to form the final vector for classifiation. The fundamental difference of this appraoch from our method is that the input to SPP network is a single fixed-size image, while in our SRNN paradim the input is a sequence of image copies of different sizes. The feature pyramid algorithm \cite{featurepyramid2016lin} is another mutli-scale neural network mainly targeted at object detection. Different from the proposed approach, the multi-scale hierarchy is created by different stages of the CNN network with spatial pooling and feature plane expansion. The input to feature pyramid network is still a single fixed-size image, rather than a sequence of image copies at different scales. 

More similar to the proposed approaches are the algorithms of LAP-GAN\cite{lapgan2015denton} and feed-forward style synthesis\cite{feedforwardstyle2016Ulyanov}. The LAP-GAN algorithm attempts to generate high-quality and fine-detailed synthetic image by using a multi-scale Lapylasian Pyramid, where separate GANs are trained for each pyramid level with a coarse-to-fine manner. This coarse-to-fine sequential generation process is similar to unrolled RNN processing. However, the main difference is that the GANs at different levels do not share the same set of parameters, and we may not be able to frame the scale propagation process under the RNN framework even the paramters are shared. The feedforward style synthesis algorithm \cite{feedforwardstyle2016Ulyanov} is somewhat similar to LAP-GAN, where multiple noise images at different scales are fed into separate network branches and eventually are combined to generate the final synthetic images. Again, the parameters for those branches are not shared, although it is also a coarse-to-fine processing paradigm.

Outside the comptuer vision and image processing research area, recurrent neural etworks have been widely in  Natual Language Processing (NLP), with improved algorithms, such as LSTM \cite{lstm1997hochreiter} and more recently GRU \cite{gru2014cho}. The proposed algorithm is partly inspired by the RNN models in NLP, where the inputs are typically variable-sized, for instance, English sentences. In NLP domain, multi-scale processing has also been explored. For example, the hierarchical multi-scale RNN (HM-RNN) \cite{HMRNN2016chung} is developed for capturing latent hierarchical structure in a sequence. However, the main difference from the proposed work in this paper is that the HM-RNN consists of RNNs defined at different scales, where each RNN still has recurrence relation along the temporal direction, whereas the recurrence relation of the SRNN is along the scale direction. 

Another related research is multi-dimenstional RNN (MD-RNN) \cite{mdrnn2011graves}. Similar to the proposed SRNN approach, MD-RNN is also developed for the vision and image processing applications. The difference is that the recurrence relation of MD-RNN is defined on the spatial domain with certain sequence ordering, whereas the recurrence relation of the proposed SRNN is on the scale dimension. 

\section{Scale Recurrent Neural Network}

The scale recurrent neural network (SRNN) is a recurrent neural network with its recurrent relation defined along the scale direction. For the image and vision applications, the scale is the spatial scale of the image. The input of a SRNN is a sequence of image copies resized from a single input image for processing. In another perspective, the input to the SRNN is an image pyramid with variable heights. If the resizing algorithm uses Gaussian kernels, the input to SRNN is a Gaussian pyramid with variable heights. 

We develop two types of SRNNs for image recognition: vanilla SRNN based on the vanilla RNN, and SRNN with half-GRU, which is an RNN with a variant of GRU. Those two types of SRNNs are described in the following subsections.

\begin{figure}[t!]
\begin{center}
\includegraphics[width=1.0\linewidth]
                   {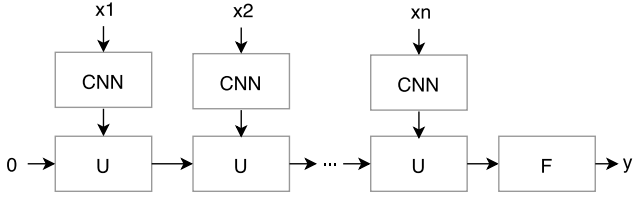}
\end{center}
\caption{Illustration of the rollout version of the SRNN (activation function is not shown)}
\label{fig:vanilla_srnn}
\end{figure}

\subsection{Vanilla SRNN}

The vanilla SRNN is just the vanilla RNN defined on image scale. For image recognition application, we define the vanilla SRNN as the following equation:
\begin{align}
\notag& h_0 = 0 & \label{eq:vanilla_srnn} \\
\notag& h_s = ReLU[ C_{NN}(x_s) + Uh_{s-1} ], & s=1,2,...n \\
& y = Softmax[F(h_n)]  &
\end{align}

Where $s$ is the scale index, and $s=1$ corresponds to the smallest scale. $x_s$ is the input image copy with scale index $s$. The input image copies are generated by resizing the input image to different scales. $C_{NN}$ is the base CNN network without the final classifier (fully connected layer). The global average pooling is used at the end of the base CNN so that for any input image size, the output is a fixed-size feature vector. $h_s$ is the hidden state at $s$ scale. $U$ is the state transition matrix. $ReLU$  is the ReLU activation function. $F$ is a fully connected layer as a classifier. $Softmax$ is the Softmax activation function which converts the output from the fully connected layer to a probability vector. 

It can be observed that there are a couple of distinctions of the above SRNN definition from the regular RNN definition. First, the recurrence relation is defined on the scale index $s$, whereas the regular RNN is defined on time index. Second, the CNN base network is used to convert the input $x_s$ into a state vector.  The CNN base network is a non-linear transform, while in the regular RNN, the transforms for the inputs are usually linear. Third, the ReLU activation function is used in SRNN, whereas in regular RNN the $tanh$ activation is usually used. The ReLU activation function allows the above system to be degenerated to a regular CNN when there is only a single scale. It also makes the system close to an ensemble classifier that averages over different scales when the state transition matrix $U$ is an identity matrix.

Figure 1 illustrates the rollout version of the aboved defined SRNN. All the $U$ modules share the same set of parameters, likewise for all the $CNN$ modules. So the SRNN can be considered as a neural network defined on an image pyramid, where each input $x_s$ corresponds to a level on the image pyramid. The use of RNN indicates that all the processing module for different pyramid layers share the same parameters, which makes it suitable for variable scale image recognition. The SRNN model can also be considered as a scale integration algorithm that combines the information from different scales.

\subsection{SRNN with half-GRU}

For image recognition application, for some object cateogries (e.g.small objects), it may be more appropriate to focus on the large-scale images while skipping some features from the small-scale images. The vanilla SRNN described above lacks the mechanism to skip or weight the features coming from previous scales. In order to achieve inter-scale feature re-weighting, a gating mechanism that is similar to the Gated Recurrent Unit(GRU) \cite{gru2014cho} may be introduced into the vanilla SRNN. However, the use of two gating functions (\textit{update} and \textit{reset} gate) in regular GRU may be an overkill for SRNN. Therefore, we use a simplied version that only uses one gating function. The resulting modified GRU is named \textit{half-GRU}. The half-GRU is similar to other simplied GRU variants developed more recently, such as those in \cite{gruvariants2017dey} and \cite{gruminimal2017heck}. 

The SRNN with half-GRU is defined mathematically as the following:

\begin{align}
\notag& h_0 = 0 & \label{eq:hgru_srnn} \\
\notag& h'_s = ReLU[ C_{NN}(x_s) + Uh_{s-1} ], & s=1,2,...n \\
\notag& z_s = Sigmoid[ W_zC_{NN}(x_s) + U_zh_{s-1}  & \\
\notag& h_s = (1-z_s){\odot}h_{s-1} + z_s{\odot}h'_s  &\\
& y = Softmax[F(h_n)] 
\end{align}

Where $h'_s $ is the hidden state estimated using the input at scale $s$ and previous hidden state $h_{s-1}$. $z_s$ is a gating function that allows to \textit{forget} some of the features from the previous scales. It uses a Sigmoid activation function, so that the resulting values range from 0 to 1. $h_s$ is the updated hidden state that combines the currently estimated state and the previous hidden state. And ${\odot}$ is element-wise multiplication operator, which allows some of the features from the previous scale to have less contribution to the final feature vector. 

\begin{figure}[t!]
\begin{center}
\includegraphics[width=1.0\linewidth]
                   {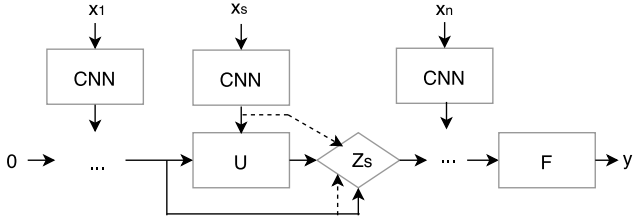}
\end{center}
\caption{Illustration of the rollout version of the SRNN with half-GRU (dashed line is the data flow for estimating the gating function)}
\label{fig:hgru_srnn}
\end{figure}

The rollout version of the SRNN with half-GRU is illustrated in Fig \ref{fig:hgru_srnn}. It can be observed that the main difference of this variant is the added gating function to weight the current state of the previous state for the next stage of processing. 

\section{Experiments}

The experiments are intended to evaluate the effectiveness of the proposed SRNN model, namely whether or not it is able to improve accuracy compared to the base CNN model. We would also like to compare its accuracy with the baseline algorithm, which is the scale average ensemble classifier.

The experiments are conducted on ImageNet-1K dataset (a.k.a ILSVRC 2012 image classification dataset) to evaluate the top-1 and top-5 image classification accuracy. Details of this dataset can be found in \cite{ilsvrc2015olga}. The same as prior work, the validation dataset is used as a proxy to test set for accuracy evaluation. Previous papers, for instance \cite{resnet2016he}, have shown that the cross-experiment variation of test accuracy is very small for the ImageNet-1K dataset due to its large size, compared with other smaller-sized datasets. 

The proposed SRNN models are implemented on PyTorch platform by modifying the ImageNet classification codebase provided by PyTorch as sample code.  

The ResNet-18 is used as the base network for comparison and evaluation.  The learning process uses a fine-tuning based training. The base CNN is initialized with the pretrained ResNet-18 model, and the final fully connected layer is initialized with the final fully connected layer in the pretrained ResNet-18 model. The state transition matrix $U$ is initialized with the identify matrix. And in the half-GRU SRNN the weight matrices for the gating function are initialized with zero matrix.

Similar to previous CNN training algorithms, the SGD with momentum is used as the learning optimizer. The momentum parameter is set to 0.9, and \textit{nesterov} momentum is enabled. The multi-step learning rate schedule is used, and the initial learning rate is set to 0.001. The learning rate is decayed by ten for every 30 epochs (i.e., times 0.1 per 30 epochs). And the traning process runs for 65 epochs for both vanilla and half-GRU SRNN. Similar to many previous CNN based classifiers, the regularization parameter, namely \textit{weight decay}, is set to $1.0\mathrm{e}{-4}$.

The data augmentation process uses the default process implemented in the PyTorch ImageNet example, which performs randomized crop, color jittering, and horizontal flips to the training images. 

Two scales are used for SRNNs for resizing the input images: one is the native image size to ResNet-18, which is $224\times{224}$, the other is $448\times{448}$. The bicubic interpolation algorithm is used for resizing the input image.

\begin{table}[b!]
\begin{center}
\begin{tabular}{l|c}
\hline
Method & Top-1 Error  \\
\hline\hline
scale=0 (224)  & 30.36  \\
scale=1 (448)  & 31.23  \\
scale ensemble (prob.)  & 29.08  \\
scale ensemble (logit)  & 28.26  \\
vanilla SRNN  & 26.56  \\
\textbf{half-GRU SRNN}  & \textbf{26.19}  \\
\hline
\end{tabular}
\end{center}
\caption{Top-1 error comparison with baseline algorithms}
\label{tb:exp_basline}
\end{table}

\subsection{Results and Comparison to Baselines}

To evaluate the effectiveness and advantages of the SRNN models, the classification results by the SRNN models are compared with the baseline algorithms, which are the average ensemble classifiers. The scale average ensemble classifier performs averaging to the outputs resulted from performing the base CNN to the image copies with different scales, which is similar to other ensemble approaches, such as 10-crop evaluation and model ensemble. Two types of scale ensemble methods are compared: ensemble over the logit outputs, which does not use the final Softmax activation, and ensemble over the probability outputs that involves Softmax activation. 

Table  \ref{tb:exp_basline} lists the top-1 error results of the proposed SRNN models, as well as the baseline algorithms (ensemble of logits and probabilities). As a reference, the results of using single-scale CNN evaluation are also listed, which include two scales: size of $224\times{224}$ and size of $448\times{448}$. The single-scale evaluation result with the size of $224\times{224}$ is roughly the same as the result reported in \cite{resnet2016he}.

It can be observed in the listed results that scale ensemble can significantly improve the classification accuracy, even if the single-scale evaluation result on the $448\times{448}$ images is actually worse than the result by using the native $224\times{224}$ size. This proves the effectiveness of the scale ensemble method for multi-scale integration. However, it is more interesting to observe that the SRNN models are able to achieve significantly lower top-1 error rate on top of the scale ensemble methods, although the computational cost introduced by adding SRNNs are marginal, since SRNN models are based on fully-connected layers rather than spatial convolution layers. The accuracy gain of the vanilla SRNN compared to the best scale ensemble model (logit ensemble) is 1.7\%, which is a significant accuracy improvement. Moreover, it can be observed that the half-GRU SRNN achieves another 0.37\% gain on top of the vanilla SRNN model. This results in a 2.07\% top-1 error reduction compared to the best scale ensemble method.

\begin{figure}[t!]
\begin{center}
\includegraphics[width=1.0\linewidth]
                   {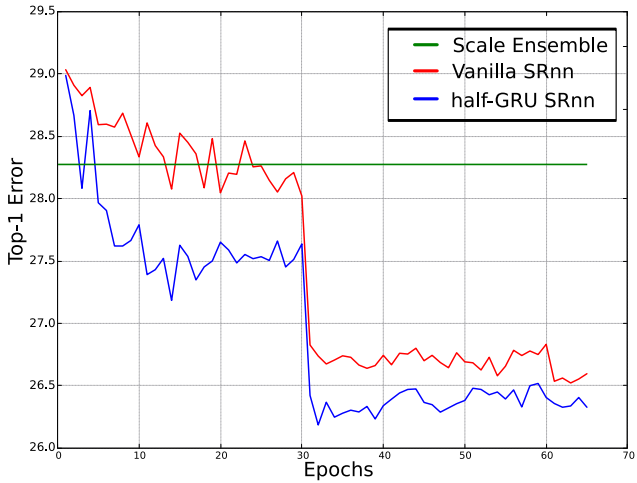}
\end{center}
   \caption{Comparison of training profiles on ImageNet-1K dataset. Scale ensemble is the average of the logit outputs }
\label{fig:train_profile}
\end{figure}

Figure \ref{fig:train_profile} also shows the training profiles of the vanilla SRNN and half-GRU SRNN models. The Y-axis of the training profiles is the top-1 error rate evaluated on the validation data set. It can be observed that the half-GRU SRNN converges faster than the vanilla SRNN, and also ends up with lower error rate.

\section{Conclusion}
We propose a new RNN-based neural network architecture for image recognition, named Scale Recurrent Neural Network (SRNN), which is distinct from the traditional feedforward CNN based framework. The new architecture is partly inspired by the human visual system, which involves both feedforward and recurrent processing. We develop two variants of SRNN: one based on vanilla RNN, and the other based on an RNN variant called half-GRU. Our experiments show that SRNN models can significantly outperform the base CNN model. Furthermore, the experiments show that the SRNN models can significantly outperform the scale ensemble based method, which performs scale integration of the base CNN model applied to different scales of the input image. The large accuracy improvement shows that the SRNN model is able to discover optimal information integration from different image scales. Finally, we also show that the SRNN with the half-GRU module can further improve the accuracy of the vanilla SRNN model through the exploitation of feature weighting across different scales.  Although SRNN is developed for the image recognition application, it can be potentially used for other applications, such as GAN or style synthesis. 

\small
\bibliographystyle{ieee}
\bibliography{egpaper_final.bib}

\end{document}